\documentclass[a4paper]{article}

\usepackage[british]{babel}
\usepackage{hyperref}
\usepackage{url}

\usepackage{graphicx}
\usepackage{tabularx,tabulary}
\usepackage[format=hang,skip=9pt]{caption}

\usepackage{lmodern}

\title{Constraint solvers: An empirical evaluation of design decisions}

\author{Lars Kotthof\/f}

\newcommand{\ecl}{ECLiPSe}
\begin{document}

\maketitle

\begin{abstract}
This paper presents an evaluation of the design decisions made in four
state-of-the-art constraint solvers; Choco, \ecl{}, Gecode, and Minion. To
assess the impact of design decisions, instances of the five problem classes
$n$-Queens, Golomb Ruler, Magic Square, Social Golfers, and Balanced Incomplete
Block Design are modelled and solved with each solver. The results of the
experiments are not meant to give an indication of the performance of a solver,
but rather investigate what influence the choice of algorithms and data
structures has.

The analysis of the impact of the design decisions focuses on the different ways
of memory management, behaviour with increasing problem size, and specialised
algorithms for specific types of variables. It also briefly considers other,
less significant decisions.
\end{abstract}

\section{Introduction}

Contemporary constraint solvers are very complex software systems. Each one of
the many available today has its own characteristics, its own design decisions
that the implementers made, and its own philosophy. The traits of a solver which
will affect the performance for a particular problem class or instance often
cannot be determined easily. Picking a particular solver is therefore a
difficult task which requires specialist knowledge about each solver and is
likely to have a significant impact on performance. On top of that, each solver
has different ways of modelling problems. Not only do users need experience with
a particular solver to model a problem in a way that enables it to be solved
efficiently, but it is also hard to objectively compare solvers.

This paper studies a small selection of constraint solvers and assesses their
performance on problem models which were made as similar as possible.

\section{Background}

The first constraint solvers were implemented as constraint logic programming
environments in logic programming languages such as Prolog in the early 1980s.
The logic programming paradigm lends itself naturally to solving constraint
problems because things like depth-first backtracking search and nondeterminism
are already built into the host language. Related ideas also arose in operations
and artificial intelligence research.

Notable developments of that time include extensions to Prolog and the CHIP
constraint programming system.

Starting in the 1990s, constraint programming found its way to procedural and
object-oriented languages, most notably C++. ILog Solver pioneered this area. It
became apparent that it would be beneficial to separate the solving of
constraint problems into two phases; modelling the problem and programming
search.

Since then, constraint solvers have improved significantly in terms of
performance as well as in terms of ease of use.

For more detailed information on the history and background of each solver, see
e.g.~\cite{cphandbook}.

\section{Surveyed constraint solvers}

The constraint solvers chosen for this paper are Choco~\cite{chocoman}, version
2.0.0.3, \ecl{}~\cite{eclipseman}, version 6.0\_42, Gecode~\cite{gecodeman},
version 2.2.0, and Minion~\cite{minion}\cite{minionman}, version 0.7. The
solvers were chosen because all of them are currently under active development.
Furthermore they are Open Source; implementation details not described in papers
or the manual can be investigated by looking at the source code.

Table~\ref{tab:solvsumm} presents a brief summary of the solvers and their basic
characteristics.

\begin{table}[htbp]
\centering
\begin{tabularx}{\linewidth}{XXXX}
\hline
\bfseries solver & \bfseries language & \bfseries year
& \bfseries modelling\\\hline
Choco & Java & 1999 & library\\
\ecl{} & C/Prolog & 1990 & library\\
Gecode & C++ & 2005 & library\\
Minion & C++ & 2006 & input file\\
\hline
\end{tabularx}
\caption{Summary of the characteristics of the investigated solvers.}
\label{tab:solvsumm}
\end{table}

\begin{description}
\item[Choco]
Choco was initially developed in the CLAIRE programming language as a national
effort of French researchers for an open constraint solver for teaching and
research purposes. Since then, it has been reimplemented in the Java programming
language and gone through a series of other changes. Version 2 is a major
refactoring to provide a better separation between modelling and solving a
problem, as well as performance improvements.
\item[\ecl]
\ecl{} is one of the oldest constraint programming environment which is still
used and in active development. It was initially developed at the European
Computer-Industry Research Centre in Munich, and then at IC-Parc, Imperial
College in London until the end of 2005, when it became Open Source. Being
implemented in Prolog, its intrinsic performance is not as high as comparable
systems implemented in other programming languages, but it is easier to specify
problems and implement new algorithms.
\item[Gecode]
``Gecode is an open, free, portable, accessible, and efficient environment for
developing constraint-based systems and applications in research, industry, and
education. Particularly important for its design is simplicity and
accessibility. Simplicity has been the key reason why Gecode is efficient and
successfully exploits today's commodity parallel hardware. Accessibility is due
to its complete reference documentation, growing tutorial documentation, and
academic publications in conferences and journals presenting key design
decisions and techniques.''\footnote{Personal communication with Christian
Schulte.}
\item[Minion]
Minion was implemented to be a solver which only requires an input file to run
and no written code. This way the solver could be made fast by not being
extensible or programmable and fixing the design decisions. It also makes it
easier to use because users do not have to write code.
\end{description}

\section{Surveyed constraint problems}

The classes of constraint problems investigated are the $n$-Queens, Golomb
Ruler, Magic Square, Social Golfers, and Balanced Incomplete Block Design
problems. The characteristics of the problems are~\cite{csplib}:-
\begin{description}
\item[$n$-Queens]
    Place $n$ queens on an $n\times n$ chessboard such that no queen is
    attacking another queen.
\item[Golomb Ruler] (CSPLib problem 6)\\
    A Golomb ruler may be defined as a set of $m$ integers
    $0 = a_1 < a_2 < \ldots < a_m$
    such that the $\frac{m(m-1)}{2}$ differences
    $a_j - a_i, 1 \leq i < j \leq m$ are distinct. Such a ruler is said to
    contain $m$ marks and is of length $a_m$. The length is to be minimised.
\item[Magic Square] (CSPLib problem 19)\\
    An order $n$ magic square is a $n\times n$ matrix containing the numbers 1
    to $n^2$, with each row, column and main diagonal equal the same sum
    $\frac{n(n^2+1)}{2}$.
\item[Social Golfers] (CSPLib problem 10)\\
    In a golf club where $m$ groups of $n$ golfers play over $p$ weeks, schedule
    the groups such that no golfer plays in the same group as any other golfer
    twice.
\item[Balanced Incomplete Block Design] (CSPLib problem 28)\\
    A Balanced Incomplete Block Design (BIBD) is defined as an arrangement of
    $v$ distinct objects into $b$ blocks such that each block contains exactly
    $k$ distinct objects, each object occurs in exactly $r$ different blocks,
    and every two distinct objects occur together in exactly $\lambda$ blocks.
    The parameters $b$ and $r$ can be derived from the other ones.
\end{description}

The choices cover a variety of different constraint problems, such as
optimisation problems and problems usually modelled with integer and Boolean
variable domains. The models involve binary constraints as well as global
constraints.

For each problem class, several different instances were chosen. This choice was
purely based on the CPU time of the models to be able to compare both long and
short runs. The instances selected were:-
\begin{description}
\item[$n$-Queens] $n = \{20,21,22,23,24,25,26,27,28,29\}$
\item[Golomb Ruler] $m = \{9,10,11,12,13\}$
\item[Magic Square] $n = \{4,5,6\}$
\item[Social Golfers] $\left<p,m,n\right> =
    \{\left<2,4,4\right>,\left<2,5,4\right>,\left<2,6,4\right>,
    \left<2,7,4\right>,\left<2,8,4\right>,\\\left<2,9,4\right>,
    \left<2,10,4\right>\}$
\item[BIBD] $\left<v,k,\lambda\right> =
    \{\left<7,3,10\right>,\left<7,3,20\right>,\left<7,3,30\right>,
    \left<7,3,40\right>,\left<7,3,50\right>,\left<7,3,60\right>,\\
    \left<7,3,70\right>\}$
\end{description}

There is insufficient space to reproduce the models for all the problems;
instead, a high-level description of the model for each problem class will be
given.

The models were derived from the examples included with the distributions of the
solvers. For some solvers and some problems the example model was simply adapted
to match the models for the other solvers, in other cases the problem was
modelled from scratch.

\begin{description}
\item[$n$-Queens] The problem was modelled with $n$ variables, one for
each queen, and one auxiliary variable for each pair of rows holding the
difference of the column positions of queens in those rows to enforce the
constraint that no two queens can be on the same diagonal. An alldifferent
constraint was enforced over the $n$ decision variables.
\item[Golomb Ruler] The Golomb Ruler model had $m$ variables, one for each tick,
and one auxiliary variable for each pair of ticks to hold the difference between
them. Additional constraints determined the value of the first tick to be 0 and
enforced an increasing monotonic ordering on the ticks. An alldifferent
constraint was enforced over the auxiliary variables holding the differences
between the ticks. The optimisation constraint minimised the value of the last
tick, which is equivalent to the length $a_m$.
\item[Magic Square] There were $n\times n$ variables for the cells of
the magic square. The constraints enforced all those variables to be different
and all rows, columns, and diagonals to sum to the magic sum. Additionally, four
constraints were introduced to break some of the symmetries in the problem; the
number in the top left square has to be less than or equal to the numbers in the
other corners of the square and the top right number has to be less than or
equal to the bottom left number.
\item[Social Golfers] The model of the Social Golfers problem used a $p\times
m\times (n\cdot m)$ matrix of decision variables. The first dimension
represented the weeks, the second one the groups, and the third one the players
by group. The constraints imposed were that each player plays exactly once per
week, the sum of the players in each group is equal to the number of players per
group specified, and each pair of players meets at most once. For the last
constraint, one auxiliary variable for each pair of players by group times weeks
times groups was introduced. Additional ordering constraints were introduced to
break the symmetries among weeks, groups, and players.
\item[Balanced Incomplete Block Design] The BIBD model introduced a matrix of
$v\times b$ decision variables. The rows were constrained to sum to $r$, the
columns to $k$, and the scalar product between each pair of rows was constrained
to equal $\lambda$. For the last constraint, one auxiliary constraint per pair
of rows times $b$ was introduced. To break some of the symmetries, ordering
constraints were put on each pair of rows and each pair of columns.
\end{description}

All models except the BIBD and Social Golfers ones used variables with integer
domains. The models of BIBD and Social Golfers used Boolean variables in the
solvers which provide specialised Boolean variables; Choco, Gecode, and Minion.
For all models, static variable and value ordering heuristics were used. The
solutions the different solvers found for each problem were the same.

Table~\ref{tab:varscons} lists the number of variables, their domains, and
constraints for each problem instance. If the domains of the auxiliary variables
are different from the domains of the main variables, they are given in
parentheses. Minion does not provide a sum equals constraint; it can however be
emulated by combining a sum less than and sum greater than constraint. This
results in a higher number of constraints for Minion; it is given in
parentheses.

\begin{table}[htbp]
\centering
\setlength\tymax{.25\textwidth}
\begin{tabulary}{\textwidth}{LLLLL}
\hline
\bfseries problem & \bfseries instance & \bfseries var\-i\-ables &
\bfseries domains & \bfseries constraints\\\hline
$n$-Queens & 20 & 210 & $\{0..19\}$ ($\{-19..19\}$) & 571 (761)\\
    & 21 & 231 & $\{0..20\}$ ($\{-20..20\}$) & 631 (841)\\
    & 22 & 253 & $\{0..21\}$ ($\{-21..21\}$) & 694 (925)\\
    & 23 & 276 & $\{0..22\}$ ($\{-22..22\}$) & 760 (1013)\\
    & 24 & 300 & $\{0..23\}$ ($\{-23..23\}$) & 829 (1105)\\
    & 25 & 325 & $\{0..24\}$ ($\{-24..24\}$) & 901 (1201)\\
    & 26 & 351 & $\{0..25\}$ ($\{-25..25\}$) & 976 (1301)\\
    & 27 & 378 & $\{0..26\}$ ($\{-26..26\}$) & 1054 (1405)\\
    & 28 & 406 & $\{0..27\}$ ($\{-27..27\}$) & 1135 (1513)\\
    & 29 & 435 & $\{0..28\}$ ($\{-28..28\}$) & 1219 (1625)\\\hline
Golomb Ruler & 9 & 45 & $\{0..81\}$ & 46 (82)\\
    & 10 & 55 & $\{0..100\}$ & 56 (101)\\
    & 11 & 66 & $\{0..121\}$ & 67 (122)\\
    & 12 & 78 & $\{0..144\}$ & 79 (145)\\
    & 13 & 91 & $\{0..169\}$ & 92 (170)\\\hline
Magic Square & 4 & 16 & $\{1..16\}$ & 15 (25)\\
    & 5 & 25 & $\{1..25\}$ & 17 (29)\\
    & 6 & 36 & $\{1..36\}$ & 19 (33)\\\hline
Social Golfers & 2,4,4 & 1088 & $\{0..1\}$ & 1133 (1293)\\
    & 2,5,4 & 2100 & $\{0..1\}$ & 2161 (2401)\\
    & 2,6,4 & 3600 & $\{0..1\}$ & 3679 (4015)\\
    & 2,7,4 & 5684 & $\{0..1\}$ & 5783 (6231)\\
    & 2,8,4 & 8448 & $\{0..1\}$ & 8569 (9145)\\
    & 2,9,4 & 11988 & $\{0..1\}$ & 12133 (12853)\\
    & 2,10,4 & 16400 & $\{0..1\}$ & 16571 (17451)\\\hline
BIBD & 7,3,10 & 1960 & $\{0..1\}$ & 1643 (1741)\\
    & 7,3,20 & 3920 & $\{0..1\}$ & 3253 (3421)\\
    & 7,3,30 & 5880 & $\{0..1\}$ & 4863 (5101)\\
    & 7,3,40 & 7840 & $\{0..1\}$ & 6473 (6781)\\
    & 7,3,50 & 9800 & $\{0..1\}$ & 8083 (8461)\\
    & 7,3,60 & 11760 & $\{0..1\}$ & 9693 (10141)\\
    & 7,3,70 & 13720 & $\{0..1\}$ & 11303 (11821)\\\hline
\end{tabulary}
\caption{Number of variables and constraints for the investigated problems.}
\label{tab:varscons}
\end{table}

The purpose of this paper is to compare the solvers on equivalent models to be
able to assess how the design decisions they have made affected their
performance. The models of the problems are in no case the optimal model for the
particular solver or the particular problem. The results cannot be seen as
providing a performance comparison of the solvers in general, as for such a
comparison the models would have to be tailored to each solver to achieve the
best performance. For such a comparison, see~\cite{solvcomp}.

This paper focuses on performance in terms of processor time; other measures
such as wall clock time and memory requirements are not evaluated.

\subsection{Amount of search}

The amount of search each solver does on each problem instance is roughly
the same. This was ensured by comparing the node counts for each instance for
the solvers which provide node counts, visually inspecting the search tree for
solvers which provide visualisation tools, and manually comparing the decisions
made at each node of the search tree for smaller instances.

The node count numbers are not reported here because because they are only
meaningful in the context of also using other means to compare the amount of
search being done.

\section{Results}

The following figures show the performance of the solvers for each problem class
and instance.

All experiments were conducted on an 8-core Intel Xeon 2.66 GHz with 16 GB of
memory running CentOS Linux 5. The CPU time was measured with the \texttt{time}
command-line utility. The numbers reported as CPU time are the sum of user and
system time. The median of five runs was taken. The coefficient of
variation\footnote{The coefficient of variation is the standard deviation
divided by the mean.} was in general less than 10\%. Instances where it was
larger are discussed below.


\begin{figure}[htbp]
\includegraphics{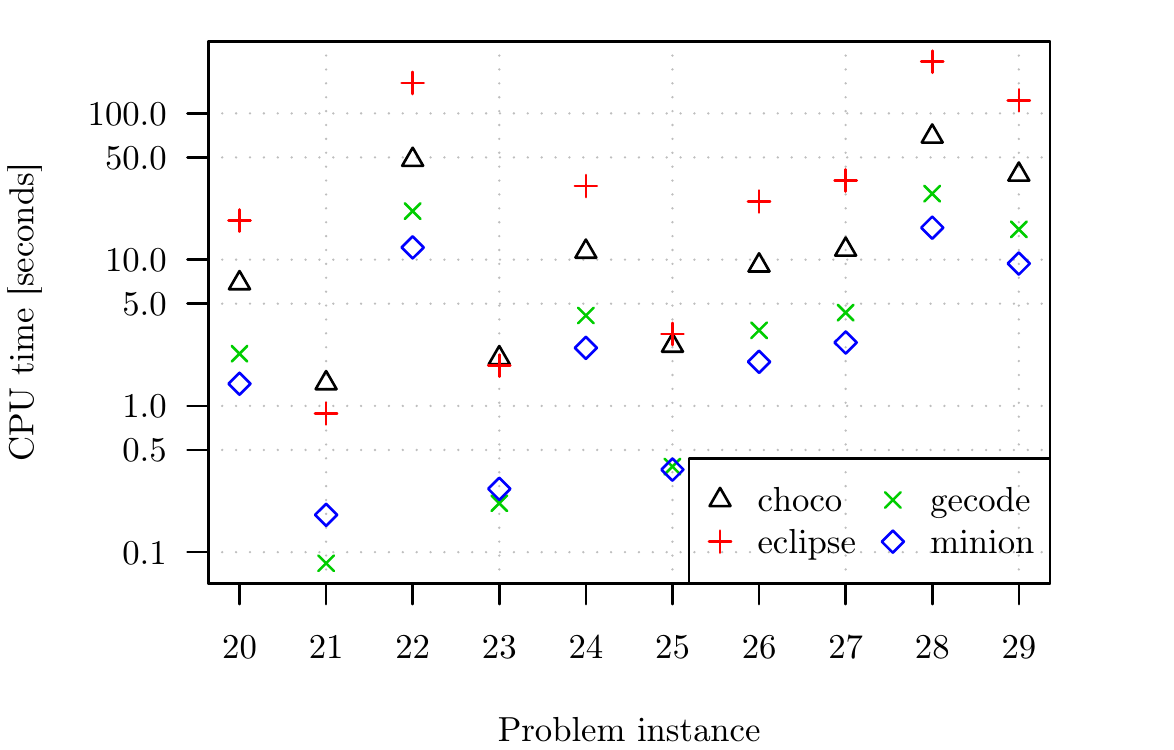}
\caption{CPU time comparison for $n$-Queens.}
\label{queens:cpu}
\end{figure}

\begin{figure}[htbp]
\includegraphics{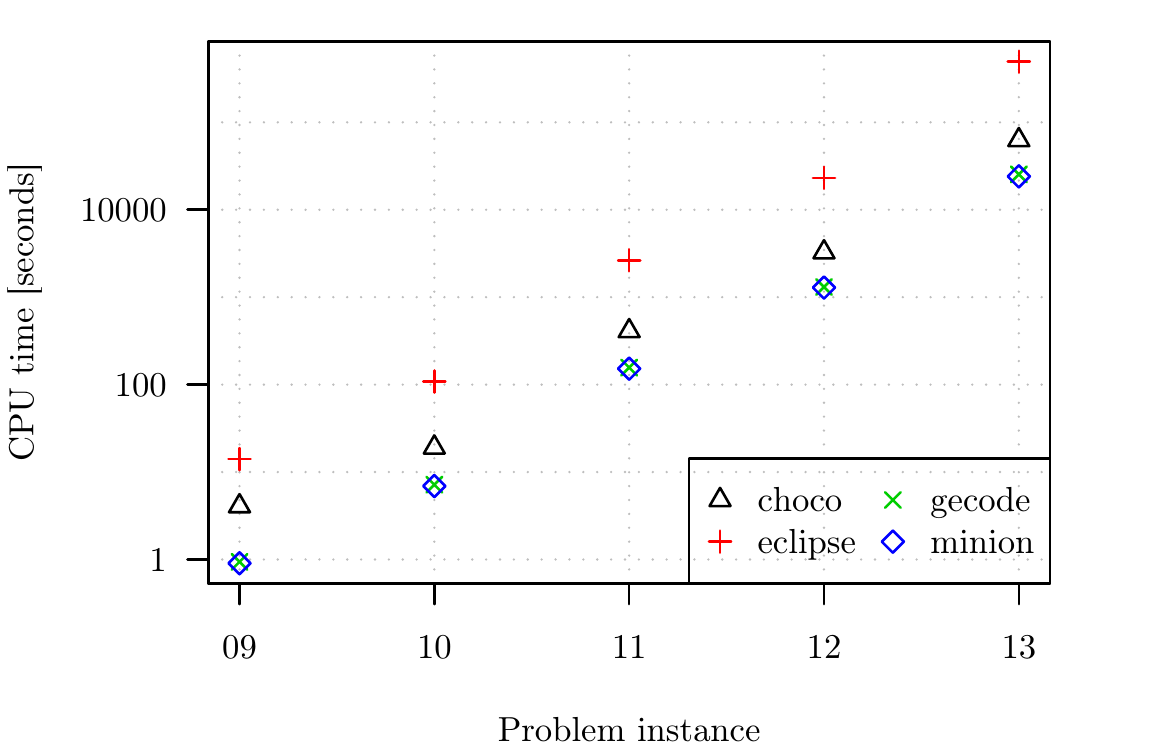}
\caption{CPU time comparison for Golomb Ruler.}
\label{golomb:cpu}
\end{figure}

\begin{figure}[htbp]
\includegraphics{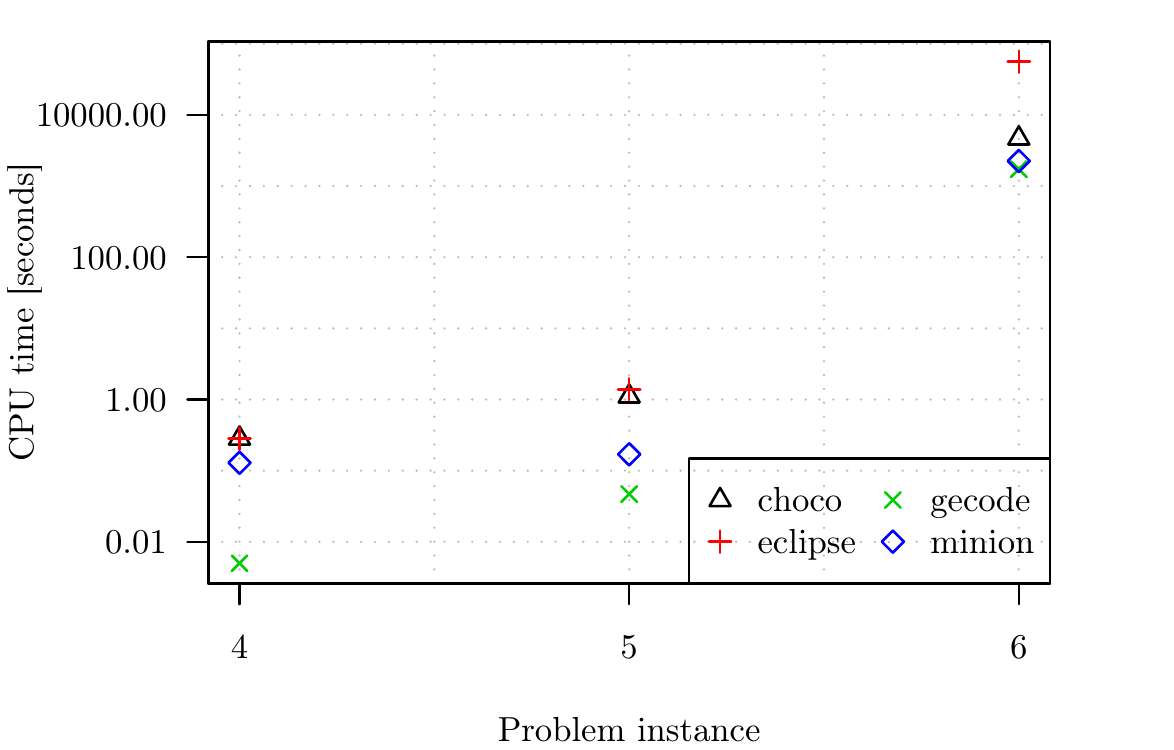}
\caption{CPU time comparison for Magic Square.}
\label{magic:cpu}
\end{figure}

\begin{figure}[htbp]
\includegraphics{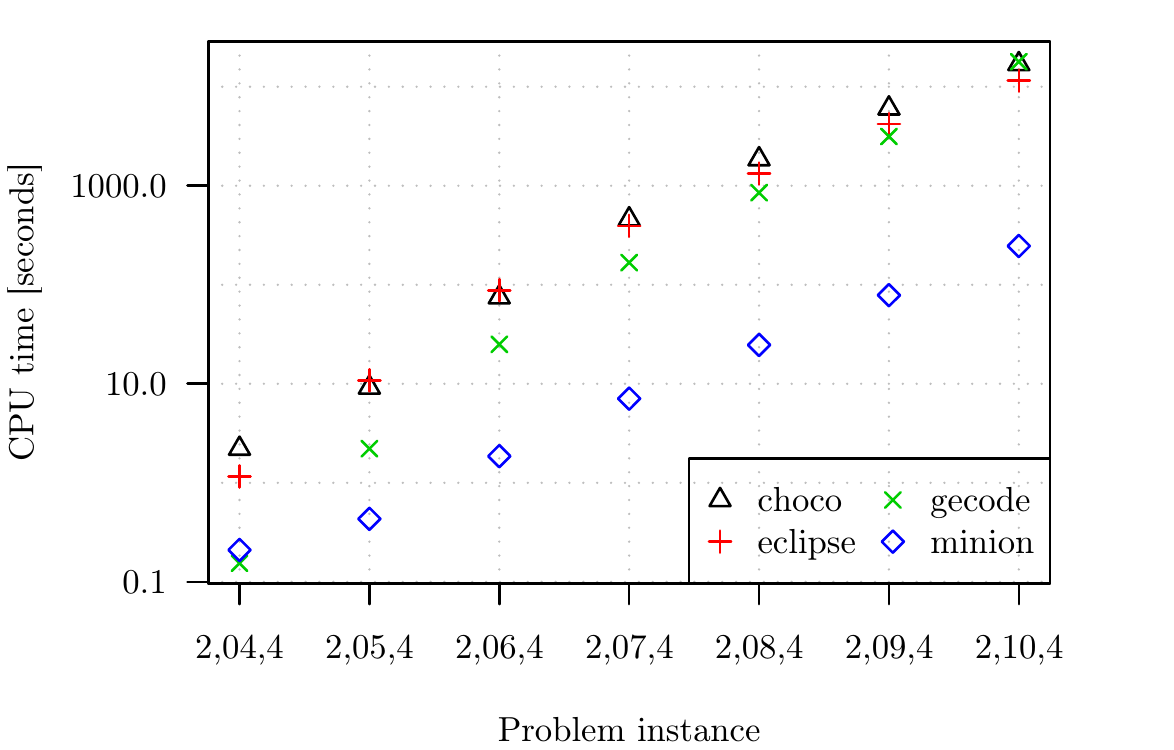}
\caption{CPU time comparison for Social Golfers.}
\label{golf:cpu}
\end{figure}

\begin{figure}[htbp]
\includegraphics{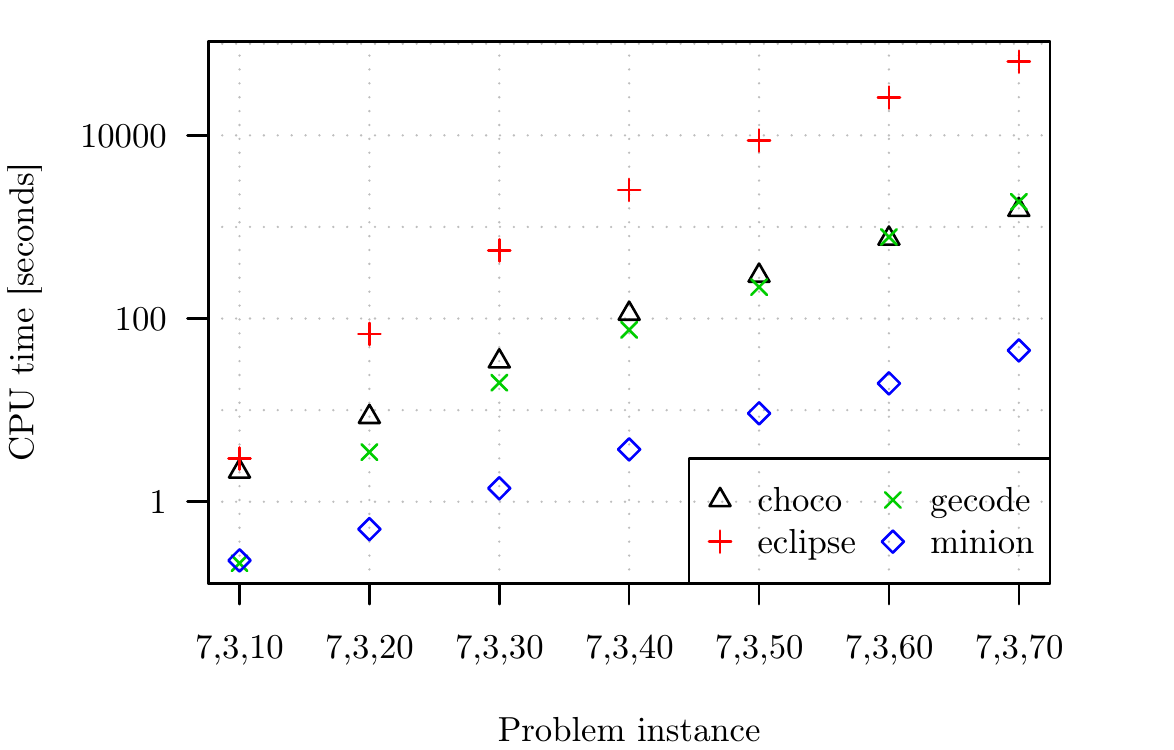}
\caption{CPU time comparison for Balanced Incomplete Block Design.}
\label{bibd:cpu}
\end{figure}

The figures show that for the Magic Square problem models, Gecode finds the
solution first. For the Golomb Ruler problem model, Gecode and Minion show a
very similar performance. For the other problem models, Minion was fastest.

\begin{figure}[htbp]
\includegraphics{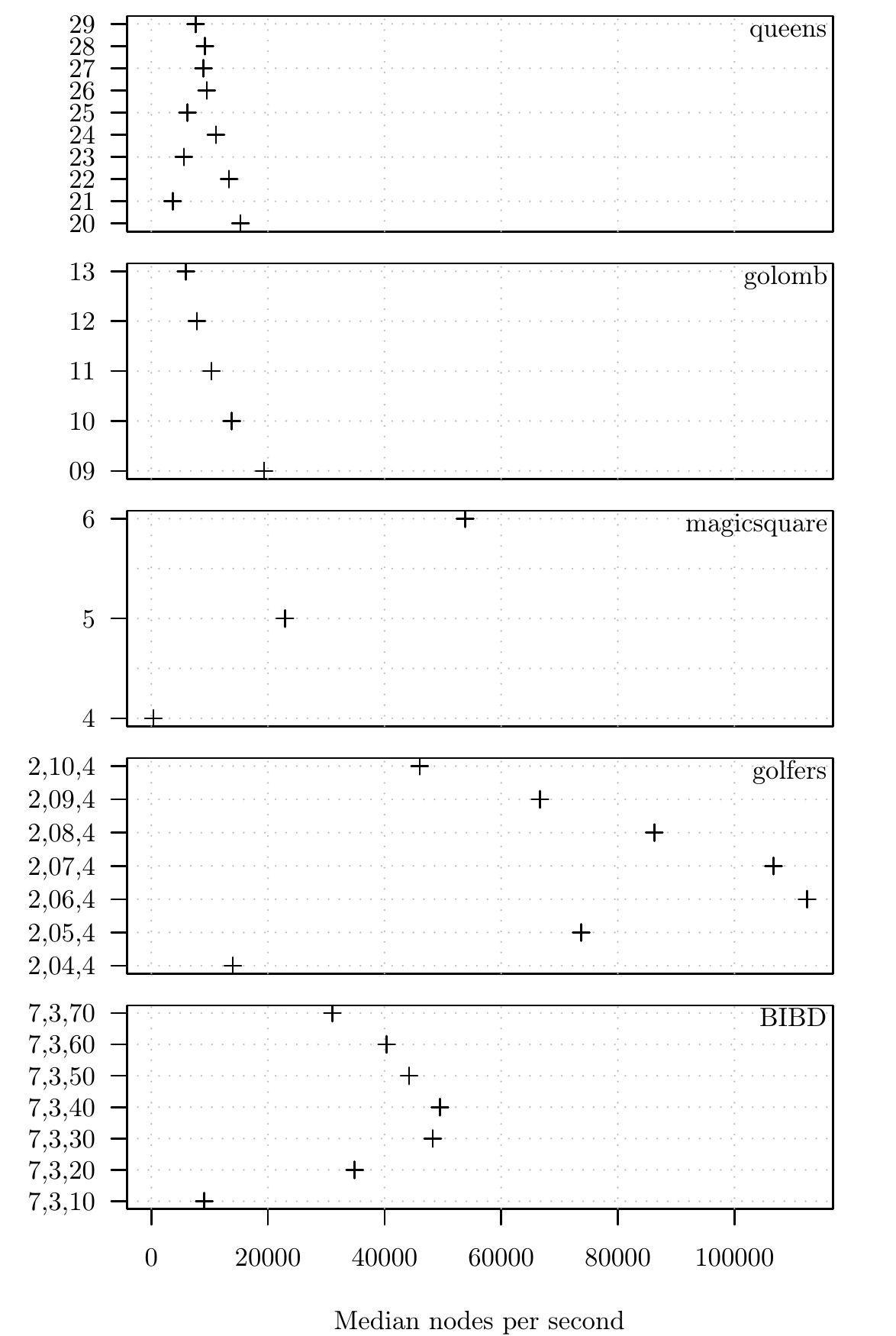}
\caption{Median nodes per second for Minion.}
\label{fig:nps}
\end{figure}

Figure~\ref{fig:nps} shows the median number of nodes per second Minion did for
each problem class and instance. A high number of nodes per second indicates
that the amount of work done at each node -- i.e.\ propagation of changes -- is
small and more search than propagation is done. For the Social Golfers and the
Balanced Incomplete Block Design problems the number of nodes per second
decreases with increasing problem size after a certain threshold. This indicates
the point where managing backtrack memory at each node becomes so expensive that
instantiating new nodes has a significant cost.

On the instances of the $n$-Queens problem, the Magic Square problem, and the
Golomb Ruler problem more propagation and less search compared to the other
problems is performed. Due to the larger domains and fewer variables, the
solvers spend a larger ratio of the total CPU time propagating changes and
revising domains than instantiating new search nodes in the tree and restoring
backtrack state.

Figure~\ref{queens:cpu} shows that the relative differences in CPU time between
the solvers stays approximately the same across different instances, save for
very small problems where the setup cost dominates the CPU time (cf.\
section~\ref{sec:setup}). The same effect is even stronger for the Golomb Ruler
problem (figure~\ref{golomb:cpu}) where the total CPU times are larger. The
gradients of the lines for the different solvers are strikingly
similar.

Figure~\ref{magic:cpu} suggests a slightly different behaviour for the Magic
Square problem; however, there are not enough data points to draw a definitive
conclusion. This problem was only run up to instances of size 6 because
instances of size 7 took too long.

These results suggest that there is no intrinsic advantage of one implementation
of propagation algorithms and data structures over another except for a constant
overhead caused by the overall implementation. They also suggest that for
problems with large variable domains the cost of propagation at each node
dominates the cost of instantiating new nodes and restoring backtrack state
regardless of the implementation of backtrack memory.

The coefficient of variation between the five runs for the 2,10,4 Social Golfers
instance for Gecode was about 20\%. Even considering the large variation, the
key point -- \ecl{} performs better than Gecode, which is roughly the same as
Choco -- remains valid.

\medskip

It is obvious from all figures that the performance differences between
different solvers can easily be several orders of magnitude. The overall
performance of a solver is affected by a variety of factors. One of them is the
programming language the solver is implemented in; others are the design
decisions made when implementing it. The following sections each look at one
of these design issues and assess its influence qualitatively and
quantitatively. There are design decisions which are not dealt with here;
however we believe that the ones addressed in this paper are the most
influential ones.

\subsection{Specialised variable implementations}

The Choco, Gecode, and Minion solvers provide specialised variable
implementations for Boolean variables. The Social Golfers and BIBD problems have
been modelled with Boolean variables and integer variables with domains
$\{0..1\}$ to assess the impact of the specialised implementation.

\ecl{} provides no variable types and uses floating point arithmetic for
everything, which gives it an inherent disadvantage over the other solvers.

\begin{figure}[htbp]
\includegraphics{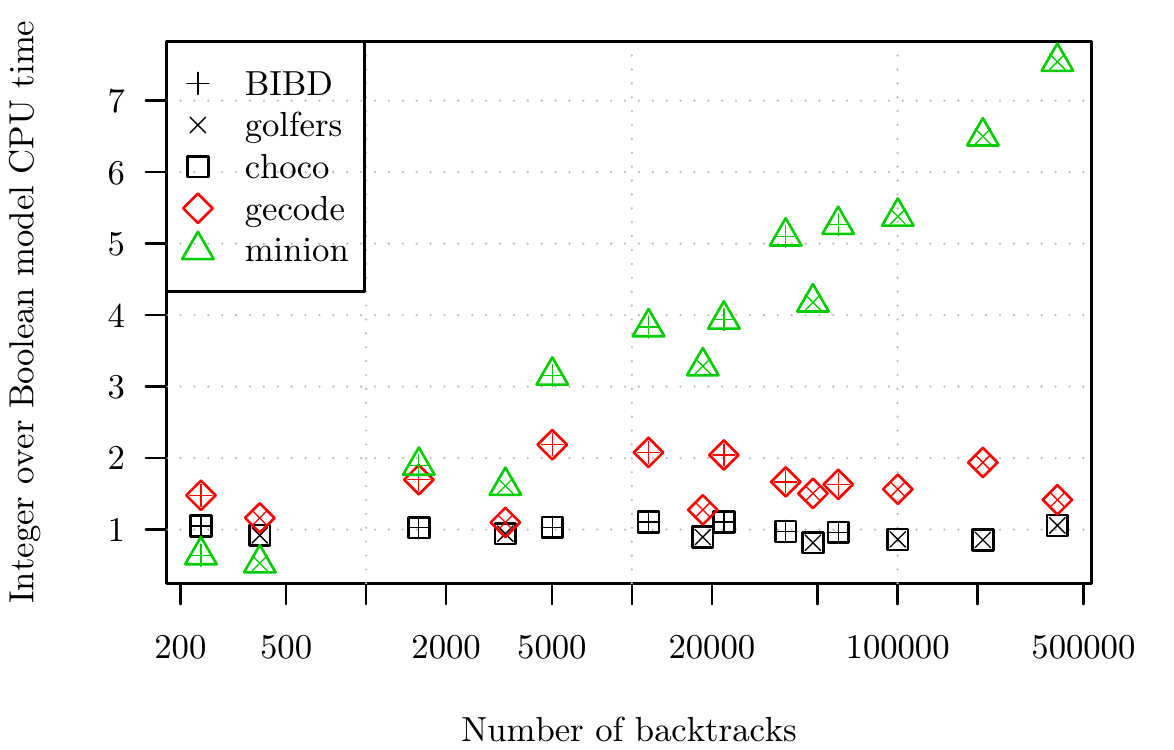}
\caption{Relative CPU time of integer model compared to Boolean model.
Shapes denote solvers, crosshairs denote problem classes. Values greater than 1
denote that the Boolean model is faster than the integer model.}
\label{fig:boolint}
\end{figure}

Figure~\ref{fig:boolint} shows the relative CPU time the model with integer
variables takes compared to the model with Boolean variables. The CPU time is
shown in relation with the number of backtracks because the correlation between
the CPU time and the number of backtracks is stronger than the correlation
between the CPU time and the number of variables~\cite{spearman}.

The results were compared with the Wilcoxon signed-rank test~\cite{wilcox}. The
differences for both problems for Gecode and Minion are statistically
significant at the 0.05 level; however the differences for Choco are not
statistically significant.

As the figure demonstrates, the specialised implementation for Boolean
variables is most effective for Minion, where the improvements over the model
with integer variables are up to more than 7 times. For Gecode the improvements
in terms of CPU time are also up to approximately 100\%. The specialised
implementation in Choco does not achieve any significant performance improvement
at all; the differences are just random noise.

The results also show that for Minion the improvement of the Boolean model over
the integer model increases as the number of backtracks increases whereas for
Gecode there is no such effect. For the smallest number of backtracks for both
the Social Golfers and the Balanced Incomplete Block Design problems, the
improvement for Choco and Gecode is larger than the improvement for Minion.

Minion provides a specialised implementation for the sum constraint, which is
used heavily in the Social Golfers and BIBD problems for variables with Boolean
domains. Gecode provides specialised Boolean implementations for all constraints
used in the models. Choco only provides a more efficient implementation of
the domain for Boolean variables.

The management of backtrack memory in Minion is slightly different for Boolean
and integer variables; for Boolean variables no trailing but only copying is
performed while for integer variables some additional trailing occurs. To assess
the effects of the specialised constraint implementation and introduction of
trailing separately, the Minion source code was modified and trailing switched
off for integer variables for the experiments described above.

The experiments were repeated with the unmodified source code of Minion. The
results showed the same picture; only the improvement of the Boolean model over
the integer model was not as significant as trailing integer variables improves
the performance slightly.

The results show that providing specialised implementations for different
variable types can achieve considerate performance improvements. The performance
improvement can be significant, as shown by Gecode and especially Minion.
It even increases for Minion as more variables are involved in the global
constraints and the size of the search tree increases.

\subsection{Setup costs and scaling}\label{sec:setup}

In all of the experiments except the Golomb Ruler problem, Gecode is the fastest
solver for the problem which takes least CPU time to solve. As the time required
to solve the problem increases, its CPU time increases more in relative terms
than that of the other solvers such that it is not the fastest solver anymore.

Both Choco and \ecl{} run in abstract machines which have to be set up when the
program starts. Minion reads an input file, parses it, and constructs the
problem to solve from that. The overhead incurred because of these issues
accounts for the differences in CPU time compared to Gecode for the small
problems. For the Golomb Ruler problem, the CPU time Gecode takes to solve the
smallest problem is equal to the time Minion takes. This is because the CPU time
required to solve this instance is large compared to the CPU required for the
smallest instances of the other problem classes -- it takes roughly a second
whereas for other problem classes the smallest instance is solved in a fraction
of a second. The overhead Minion incurs by parsing the input file accounts only
for a small fraction of the total CPU time and therefore it is as fast as
Gecode.

Figure~\ref{golf:cpu} shows that for the Social Golfers problem, \ecl{} scales
better than the other solvers with respect to the increase in CPU time with
increasing problem size. From the 2,7,4 instance, it is faster than Choco, and
for the largest instance it is faster than Gecode as well. Extrapolating past
the end of the graph, it is possible that for very large instances \ecl{} could
be faster than Minion.

Figure~\ref{bibd:cpu} on the other hand shows a different picture. Here the
relative increase in CPU time \ecl{} and Gecode require to solve the problem
as it becomes more difficult to solve is significantly larger than that of Choco
and Minion. For the 7,3,60 problem instance, Choco is faster than Gecode despite
being slower before.

Both graphs are strikingly similar when disregarding \ecl{}. For both problems,
the relative distance between the lines for Choco and Minion stays more or less
the same, whereas Gecode is about the same as Minion for the smallest problem
and about the same as Choco for the largest problem.

\begin{table}[htbp]
\centering
\begin{tabulary}{\linewidth}{JJJ}
\hline
\bfseries problem & \bfseries instance & \bfseries backtracks\\\hline
$n$-Queens & 20 & 5960\\
     & 21 & 177\\
     & 22 & 43783\\
     & 23 & 389\\
     & 24 & 7337\\
     & 25 & 606\\
     & 26 & 4922\\
     & 27 & 6465\\
     & 28 & 39467\\
     & 29 & 18687\\\hline
Social Golfers & 2,4,4 & 398\\
    & 2,5,4 & 3343\\
    & 2,6,4 & 18497\\
    & 2,7,4 & 48030\\
    & 2,8,4 & 100201\\
    & 2,9,4 & 209387\\
    & 2,10,4 & 399498\\\hline
BIBD & 7,3,10 & 239\\
    & 7,3,20 & 1579\\
    & 7,3,30 & 5019\\
    & 7,3,40 & 11559\\
    & 7,3,50 & 22199\\
    & 7,3,60 & 37939\\
    & 7,3,70 & 59779\\\hline
\end{tabulary}
\caption{Number of backtracks for \ecl{}.}
\label{tab:backtracks}
\end{table}

Both the Social Golfers and BIBD problem classes have a large number of
variables and constraints. The key difference is that on instances of the Social
Golfers problem, more backtracks are performed (cf.\
table~\ref{tab:backtracks}). This indicates that the implementation of
backtracking and restoration of previous state for problems with many variables
is implemented more efficiently in \ecl{} than in the other solvers.

The following sections look at memory management in more detail.

\subsection{Memory management}

Table~\ref{tab:memory} summarises the memory management approaches taken for
the different solvers.

\begin{table}[htbp]
\centering
\begin{tabulary}{\linewidth}{JJJ}
\hline
\bfseries solver & \bfseries backtracking approach & \bfseries garbage
collection\\\hline
Choco & trailing & yes (Java)\\
\ecl{} & trailing & yes (custom)\\
Gecode & copying/recomputation & no\\
Minion & copying/trailing & no\\\hline
\end{tabulary}
\caption{Summary of memory management approaches.}
\label{tab:memory}
\end{table}

The following sections are mostly concerned with the different ways of
implementing backtrack memory, as this is the most important memory management
decision to be made in a constraint problem solver.

\subsubsection{Recomputation versus copying}

Gecode provides parameters which can be given to the solver executable to tune
the ratio of copying vs.\ recomputation. The $n$-Queens problem, the Social
Golfers problem, and the Balanced Incomplete Block Design problem were rerun
with recomputation distances of 1 (full copying -- the same behaviour as
Minion), 8 (the default), 16, and 32. The adaptive recomputation distance was
left at the default value of 2~\cite{consservices}.

The results were compared with the Kruskal-Wallis one-way analysis of variance
test~\cite{kruskal}. The differences are not statistically significant because
of the large variation among the CPU times for the problem instances; however
when comparing the differences between doing a full copy at each node
(recomputation distance 1) and the other recomputation distances with the
Wilcoxon test the differences were statistically significant at the 0.05 level.

\begin{figure}[htbp]
\includegraphics{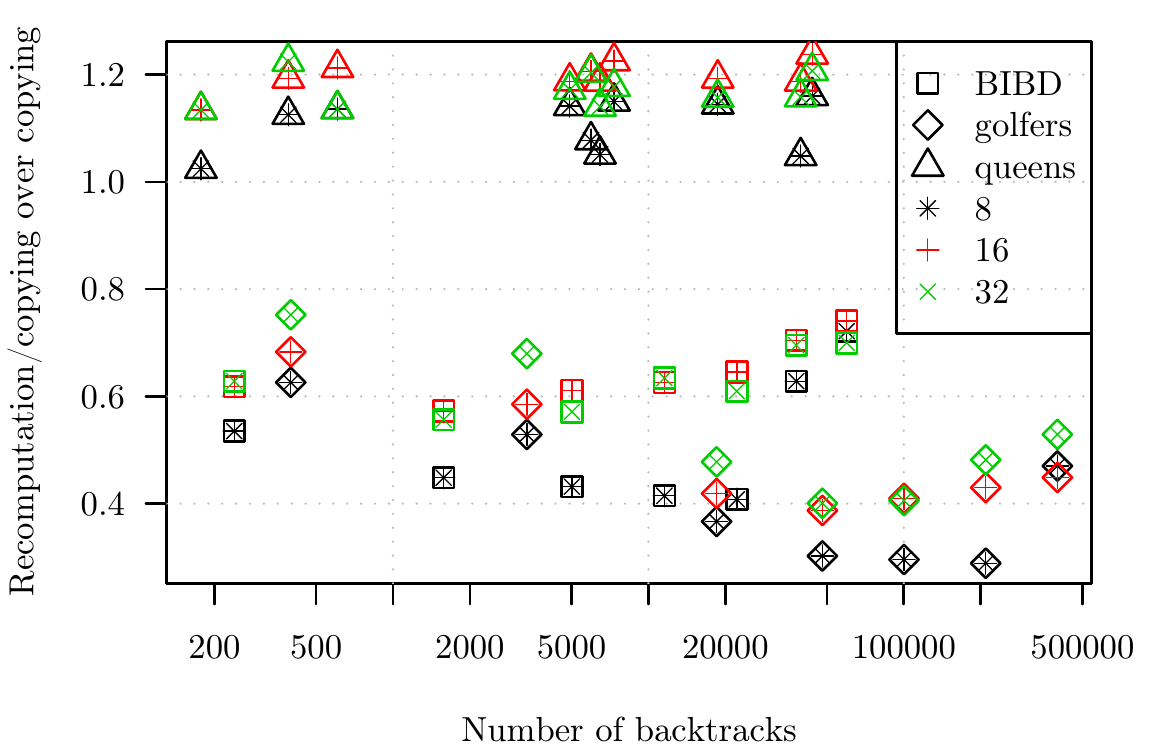}
\caption{CPU time of different levels of recomputation and copying over CPU time
of copying. Shapes denote problem classes, crosshairs denote recomputation
distances. Values less than 1 denote that copying and recomputation is faster
than copying at every node.}
\label{fig:copy}
\end{figure}

Figure~\ref{fig:copy} shows the results for all the problems and recomputation
distances. The CPU times of the runs with a recomputation distance $>1$ are
divided by the CPU times for a recomputation distance of 1. Values larger than 1
mean that doing a full copy at every node performs better than a recomputation
distance of $>1$. Note that the default recomputation distance in Gecode is 8,
i.e.\ the results shown in figures~\ref{queens:cpu}, \ref{golf:cpu},
and~\ref{bibd:cpu} are not the CPU times which the other CPU times are divided
by.

The CPU time is influenced by both the number of backtracks and the number of
variables; however for this comparison the correlation between CPU time and
number of backtracks was stronger than the correlation between CPU time and
number of variables.

For all instances and recomputation distances of the $n$-Queens problem, making
a full copy at every node of the search tree performs better than a
recomputation distance $>1$. This suggests that for problems with only few
variables it is better to always copy. The performance improvement is only up to
about 22\% though.

The Social Golfers and Balanced Incomplete Block Design problems show that for
problems with many variables, it is cheaper not to copy at every node, but to do
some recomputation. The results show that the performance improvement can be up
to about 70\% with recomputation. They also demonstrate that the optimal
recomputation distance increases as the number of backtracks increases. For both
the Social Golfers and the Balanced Incomplete Block Design, the default
recomputation distance of 8 performs best for all but the largest instance of
the respective problem, where the recomputation distance of 16 is better.

Furthermore there appears to be a problem-specific threshold in terms of number
of backtracks which marks a change in the performance improvement for copying at
every node -- before the threshold the improvement increases with increasing
number of backtracks, after the threshold it decreases.

Figure~\ref{golf:cpu} for example shows that Minion performs better than Gecode
despite full copying and no recomputation. This is because Gecode and Minion use
different implementations of copying backtrack memory. Whereas Gecode keeps a
list of pointers to objects to be copied and traverses that list, Minion
allocates everything that needs to be restored when backtracking in a continuous
memory region and simply copies the whole region. The advantage of Gecode's
approach is that a finer-grained control over the used memory is possible, but
Minion's approach wins in terms of overhead when copying at every node.

Choco provides a facility to change the backtrack strategy to both recomputation
and copying as well; however not a combination of the two. Using only
recomputation performed worse than copying by several orders of magnitude and is
therefore not considered here.

\subsubsection{Copying versus trailing}

Choco allows to change the default backtrack strategy of trailing to copying.
The $n$-Queens, the Social Golfers, and the Balanced Incomplete Block Design
problems were rerun with copying instead of trailing for backtrack memory.

The results were compared with the Wilcoxon test. The differences are
statistically significant at the 0.01 level. The CPU times are shown in relation
to the number of backtracks because the correlation between the CPU time and the
number of backtracks is stronger than the correlation between the CPU time and
the number of variables.

\begin{figure}[htbp]
\includegraphics{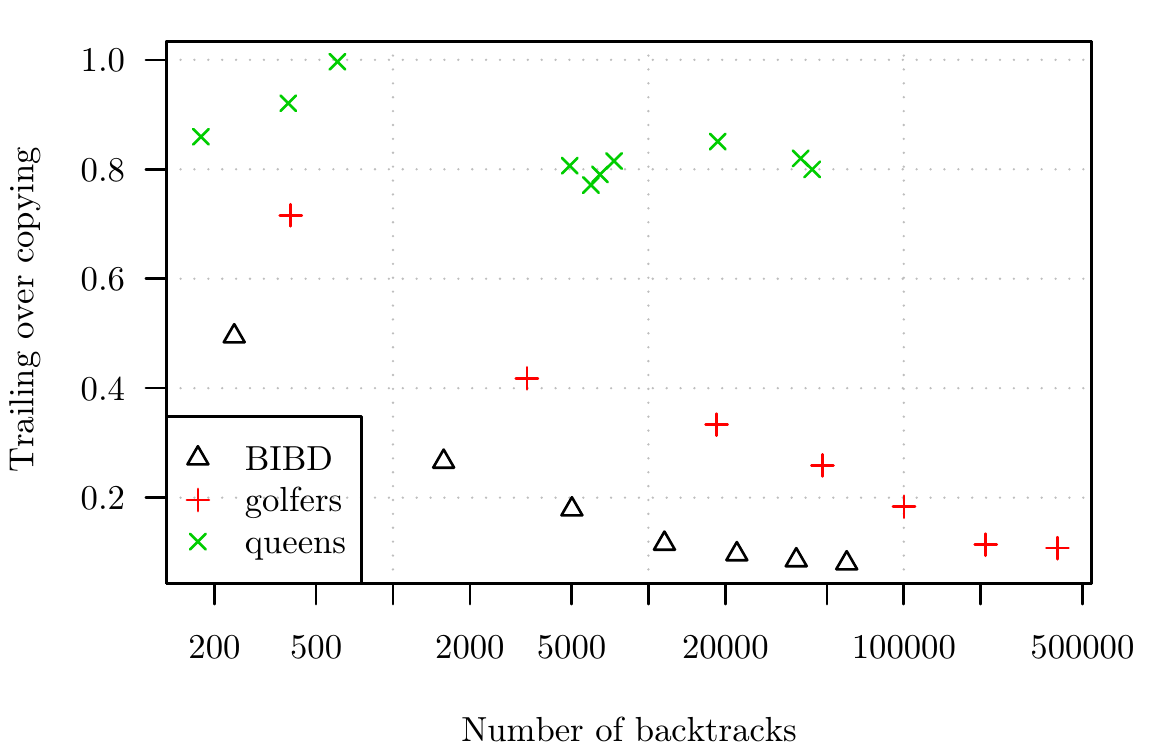}
\caption{CPU time of trailing over CPU time of copying. Values less than 1
denote that trailing is faster than copying.}
\label{fig:trail}
\end{figure}

Figure~\ref{fig:trail} shows the results. For all instances of all problems,
trailing performs better than copying. For the $n$-Queens problem the
differences are only up to about 20\% because of the small number of variables
(cf.\ table~\ref{tab:varscons}). The results for the Social Golfers and the
Balanced Incomplete Block Design problems show that the relative difference
between trailing and copying backtrack memory increases as the number of
backtracks increases.

The results suggest that trailing backtrack memory performs better than copying
backtrack memory; especially with increasing number of backtracks and variables.
This is most likely limited to Choco though; Minion for example uses a different
implementation of copying backtrack memory which scales much better with
increasing number of variables and has less overhead -- instead of copying each
variable domain individually, it only copies one contiguous memory region.

In general the results show that for problems with many backtracks trailing
backtrack memory performs better than copying backtrack memory. The following
section investigates this further.

\subsubsection{Sensitivity to number of variables}

To further assess the impact of the backtrack strategy on the overall
performance of the solvers, the $n$-Queens and the Social Golfers problems were
remodelled with more auxiliary Boolean variables. No additional constraints were
imposed on the variables to keep the amount of search the same.
Table~\ref{tab:manyvars} summarises the numbers of variables for the normal and
for the extended model.

\begin{table}[htbp]
\centering
\begin{tabulary}{\linewidth}{JJJJ}
\hline
\bfseries problem & \bfseries instance & \bfseries normal model & \bfseries extended model\\\hline
$n$-Queens & 20 & 210 & 1920\\
    & 21 & 231 & 2121\\
    & 22 & 253 & 2332\\
    & 23 & 276 & 2553\\
    & 24 & 300 & 2784\\
    & 25 & 325 & 3025\\
    & 26 & 351 & 3276\\
    & 27 & 378 & 3537\\
    & 28 & 406 & 3808\\
    & 29 & 435 & 4089\\\hline
Social Golfers & 2,4,4 & 1088 & 9728\\
    & 2,5,4 & 2100 & 19200\\
    & 2,6,4 & 3600 & 33408\\
    & 2,7,4 & 5684 & 53312\\
    & 2,8,4 & 8448 & 79872\\
    & 2,9,4 & 11988 & 114048\\
    & 2,10,4 & 16400 & 156800\\\hline
\end{tabulary}
\caption{Number of variables for normal and extended $n$-Queens and Social
Golfers models.}
\label{tab:manyvars}
\end{table}

The increased number of variables should have no or little impact on performance
for solvers which use trailed memory for backtracking as they only record the
changes to variables and the additional variables are never changed. The impact
for solver which use other types of backtrack strategies should be
considerable. Any effects caused by the different types of backtrack memory
should be much more significant for the Social Golfers problem instances than
for the $n$-Queens problem instances because of the significantly higher number
of backtracks (cf.\ table~\ref{tab:backtracks}).

The purpose of these experiments is twofold. First, to assess the influence the
implementation of backtrack memory when more variables are added, and second,
an estimation of the fraction of total CPU time which is spent managing
backtrack memory. This can be estimated from the influence of the backtrack
strategy on the total CPU time.

\begin{figure}[htbp]
\includegraphics{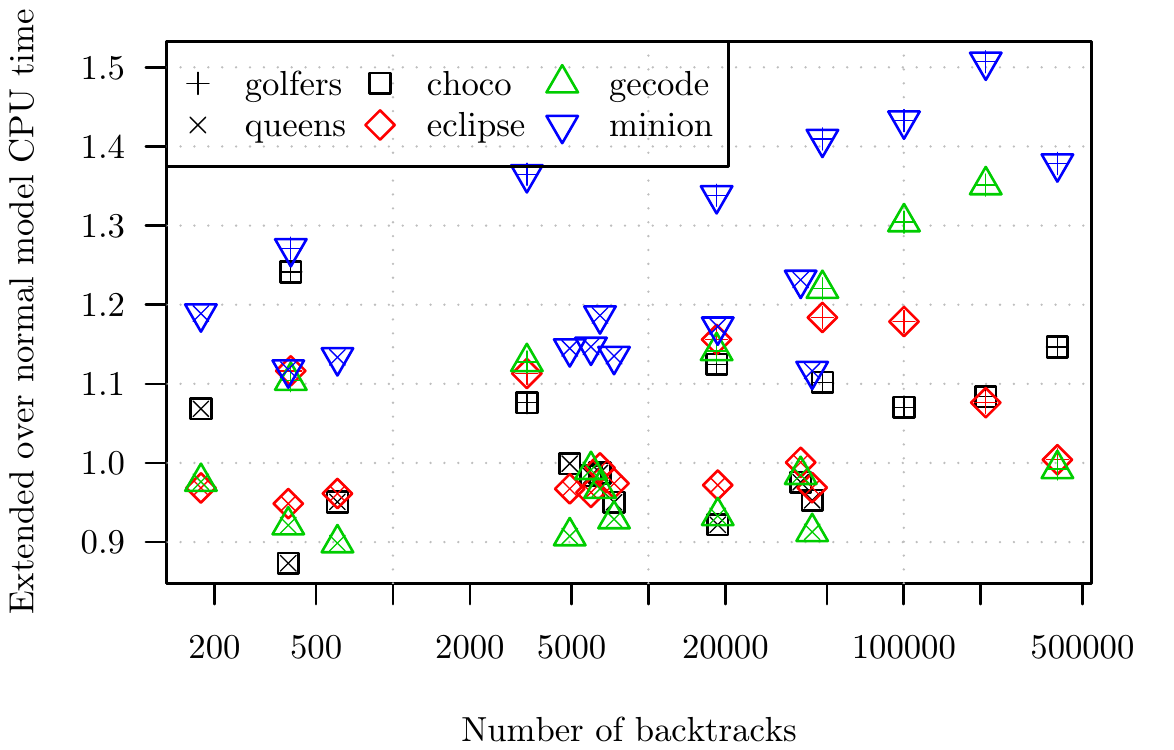}
\caption{CPU time of extended model with more variables over CPU time of normal
model. Shapes denote solvers, crosshairs denote problem classes. Values greater
than 1 denote that the CPU time for the normal model was less than the CPU time
for the extended model.}
\label{fig:manyvars}
\end{figure}

The results for each solver and problem class were compared with the Wilcoxon
test. The differences for the $n$-Queens problem are statistically significant
at the 0.05 level. For the Social Golfers problem the differences for Choco,
\ecl{}, and Minion are significant at the 0.05 level; the differences for Gecode
are not statistically significant.

Figure~\ref{fig:manyvars} shows the CPU time of the extended model over the
CPU time of the normal model. The solver which is affected most by the change is
obviously Minion; followed by Gecode. The differences of up to about 35\% for
Gecode suggest that it is affected, but the variation between individual runs
is too much to make the differences statistically significant. Choco and \ecl{}
are affected to a much lesser extent.

The results correspond exactly to the expectations. Minion takes the biggest hit
in terms of performance because it uses copying for backtrack memory and has to
copy more data. Gecode combines copying with recomputation and is therefore less
affected, as no recomputation has to be performed for the additional variables.
Both Choco and \ecl{} do trailing and are least affected by the addition of more
variables.

Figure~\ref{fig:manyvars} also shows that the CPU time of the model with more
variables increases for Gecode and Minion as the number of backtracks (and
therefore the size of the search tree) increases. For Choco and \ecl{} it stays
approximately the same. Again, this result is expected because while for copying
backtrack memory the amount of work to be done increases at each node, it stays
approximately the same for trailing backtrack memory.

The number of variables and backtracks for the $n$-Queens problem instances are
much less than for the Social Golfers instances (cf.\ tables~\ref{tab:manyvars}
and~\ref{tab:backtracks}) and therefore the expected effects are less
significant. The differences for Minion stand out from the differences for the
other solvers, suggesting that it is the only one which was truly affected by
the changes. The differences are significantly smaller than for the Social
Golfers problem though.

The results show that managing backtrack memory can account for a
significant part of the total CPU time. Adding new variables without any
constraints on them does not increase the work to be done for propagating
changes, but nevertheless the total CPU time can increase significantly.
For the remodelled $n$-Queens problem the differences for Minion are up to about
23\% even though more work is done propagating changes than exploring the search
tree (cf.\ discussion for figure~\ref{fig:nps}). For the Social Golfers problem,
which has many more variables (cf.\ table~\ref{tab:manyvars}), the proportion of
the CPU time spent on managing backtrack memory is even larger; the differences
are up to about 50\% for Minion.

\subsubsection{Garbage collection}

\ecl{} is the only solver which provides garbage collection and a facility to
switch it off. The $n$-Queens, the Social Golfers, and the Balanced Incomplete
Block Design problem classes were rerun with garbage collection turned off.

The results were compared to the results with garbage collection switched on
with the Wilcoxon test. The differences are statistically significant at the
0.01 level.

The results are shown in relation to the number of backtracks because the
correlation between CPU time and the number of backtracks is stronger than the
correlation between CPU time and the number of variables.

\begin{figure}[htbp]
\includegraphics{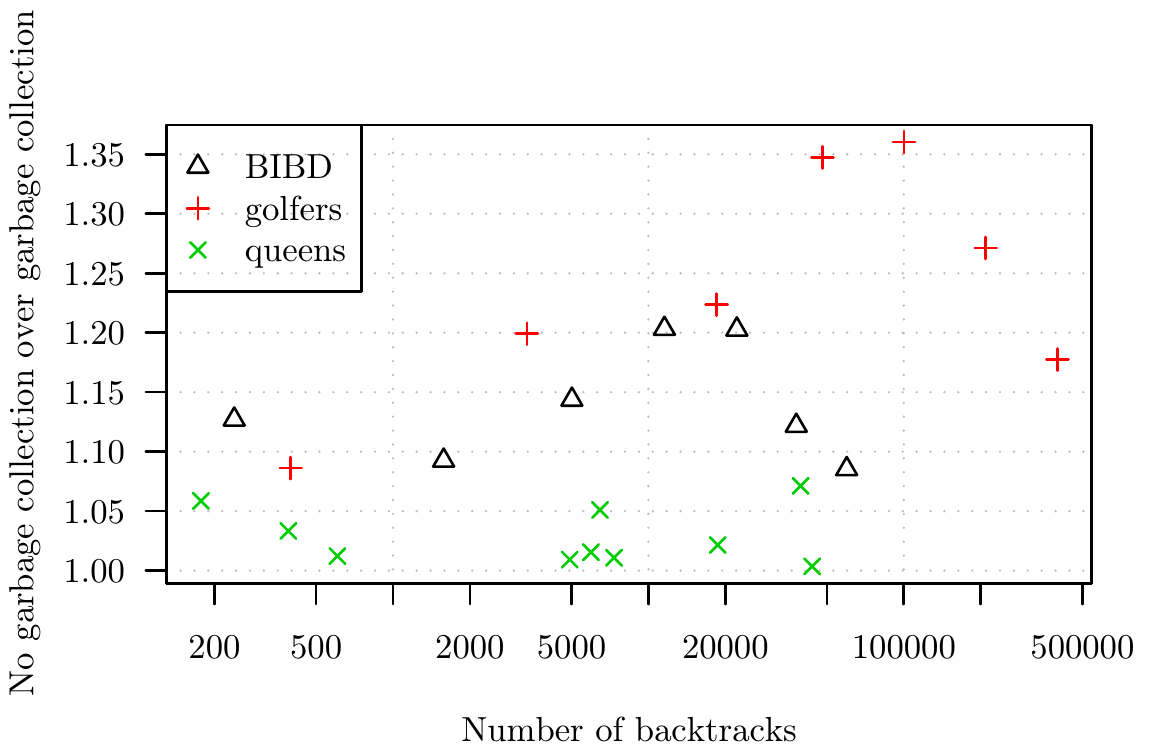}
\caption{CPU time of run with garbage collection switched off over CPU time of
run with garbage collection switched on. Values greater than 1 denote that
garbage collection is faster than no garbage collection.}
\label{fig:garbage}
\end{figure}

Figure~\ref{fig:garbage} shows that the CPU times for the runs with garbage
collection switched off are up to about 35\% higher than those with garbage
collection switched on for the Social Golfers problem. The CPU times for
Balanced Incomplete Block Design are similar, but less pronounced. A possible
reason for that is the lower number of backtracks. The amount of memory that the
$n$-Queens instances use is so small that the differences are not significant.

The results show that the benefits of garbage collection do not only include a
smaller memory footprint, but also increases in performance in terms of CPU
time. These results may not be applicable for other implementations of backtrack
memory though. The second key point which can be concluded from
figure~\ref{fig:garbage} is that while garbage collection can improve
performance up to a certain number of backtracks, the improvements become
smaller as the number of backtracks and the total amount of memory the
problem requires grows. It is conceivable that for much larger problem instances
than investigated here, the CPU time with garbage collection becomes larger than
the CPU time without garbage collection.

\subsection{Order of propagators}

Choco, \ecl{}, and Gecode provide facilities to attach priorities to
propagators, i.e.\ changes will not be propagated in the order they are made in,
but according to a priority value. Minion does not provide such a facility.

For the investigated problems, Choco does not make use of the priorities, i.e.\
the priority is the same for everything.

In \ecl{} some of the global constraints such as alldifferent and element are
processed with a higher priority than constraints of a lower arity. Only the
alldifferent constraint is used in the $n$-Queens, the Golomb Ruler, and the
Magic Square problems.

Gecode orders the propagators according the complexity of the propagation
function, which is defined when the propagator is implemented. Experiments for
some problem instances were conducted with the propagator queue reversed. For
the Balanced Incomplete Block Design problem, no differences at all were
observed, whereas for example for the $n$-Queens problem there were differences
in terms of CPU time. In all cases the maximum difference was only a small
fraction of the total CPU time though.

\subsection{Types of constraints}

Choco, \ecl{}, and Gecode offer basic constraints which can be combined into
more complex ones to a larger extent than Minion. For the particular models used
in this paper the same constraints were used and this did not have any negative
impact in terms of performance; for other applications it simplifies modelling
problems though and is therefore also likely to have an impact on performance.
For example the $n$-Queens problem could be modelled without auxiliary variables
in Choco, \ecl{}, and Gecode and is likely to be perform better than a model of
the same problem in Minion which has to use auxiliary variables.

On the other hand Minion provides an implementation of the sum constraints with
watched literals, which could improve its performance~\cite{watchlits}.

An interesting point is that Minion does not have a sum-equals constraint, but
only sum-greater-or-equal and sum-less-or-equal constraints. The semantics of
the sum-equals constraint can be achieved by combining the two constraints, but
this increases the total number of constraints; in some cases considerably.
Nevertheless there does not seem to be a negative impact on performance, on the
contrary. This indicates that the most obvious way to implement a constraint may
not always be the most efficient one.

\subsection{Optimisation problems}

The investigated solvers implement several different approaches to handling
optimisation problems; Choco and Minion handle the value to be minimised in
specialised implementations of search, \ecl{} adjusts the bounds of the cost
variable while Gecode imposes additional constraints on it.
Figure~\ref{golomb:cpu} suggests that there is no intrinsic advantage of one way
over the other.

\section{Conclusion}

We presented a comprehensive comparison and evaluation of the implementation
design decisions in state-of-the-art constraint problem solvers. The experiments
provide not only a qualitative, but also a quantitative comparison of different
implementation approaches.

The results show that choosing one design decision over another when
implementing a constraint solver does not usually give performance benefits in
general. The exception are specialised variable implementations -- implementing
specialised versions of constraints and propagators for the different variable
types improves performance significantly.

The design decisions associated with memory management, such as backtrack
memory, are much harder to classify. Depending on the problem to solve and the
number of variables and constraints involved, a particular implementation of
memory management will perform better than others. This does not only depend on
the type of problem, but also on the size of the problem though. The results do
show however that memory management can account for a significant part of the
total CPU time required to solve a problem.

The large differences among the CPU times the individual solvers take emphasise
the importance of choosing the right solver for a given task. This decision is
absolutely crucial to performance. In an ideal world, a solver would, given a
particular problem, adapt its design decisions and provide an implementation
specialised for this problem.

The performance of the individual solvers in the experiments should \emph{not}
be taken as a benchmark or as a suggestion which of these solvers to use for a
given problem. The focus of the experiments was to compare the solvers on models
which are as similar as possible. For any other application, the problem model
will be tuned for a particular solver to use its specific strengths which cannot
be compared here. It is entirely possible that with a carefully-tuned model a
solver which performs badly in an experiment reported here becomes much better
than any other solver.

\section{Acknowledgements}

The author is indebted to Ian Miguel and Ian Gent for many helpful comments on
drafts of this paper. Thanks to the \ecl{} and Gecode mailing list participants
for answering questions concerning the implementation of the respective solvers.
Thanks also go to Warwick Harvey for modelling the Social Golfers problem in
\ecl{} and Mikael Lagerkvist for pointing out a mistake in an earlier version of
this paper.

\end{document}